\documentclass[letterpaper, 10 pt, conference]{ieeeconf}  

\IEEEoverridecommandlockouts                              
\usepackage{lineno}
\usepackage{cite}
\usepackage{algorithm}
\usepackage{algpseudocode}
\usepackage{amsmath,amssymb,amsfonts}
\usepackage{graphicx} 
\usepackage{epstopdf}
\usepackage{textcomp}
\usepackage{xcolor}
\usepackage{subfigure}
\usepackage{booktabs}
\usepackage{multirow}
\usepackage{mathrsfs}
\usepackage{soul}
\usepackage{bbding}
\usepackage{setspace}
\usepackage{comment}
\usepackage{framed}
\usepackage{color}
\usepackage{soul}

\soulregister\cite7
\soulregister\ref7 
\newcommand{\tabincell}[2]{\begin{tabular}{@{}#1@{}}#2\end{tabular}}

\title{Estimating the Lateral Motion States of an Underwater Robot by Propeller Wake Sensing Using an Artificial Lateral Line}

\author{Jun Wang, Dexin Zhao, Youxi Zhao and Feitian Zhang*, Tongsheng Shen*
\thanks{J. Wang (wjun@stu.pku.edu.cn) is with the Department of Advanced Manufacturing and Robotics, College of Engineering, Peking University, Beijing, China, and also with the National Innovation Institute of Defense Technology, Beijing, China.
}%
\thanks{D. Zhao (zhaodx2008@163.com) and T. Shen (shents\_bj@126.com) are with the National Innovation Institute of Defense Technology, Beijing, China.}
\thanks{Y. Zhao (2101111921@stu.pku.edu.cn) is with the Department of Mechanics and Engineering Science, College of Engineering, Peking University, Beijing, China.}
\thanks{F. Zhang (feitian@pku.edu.cn) is with the Department of Advanced Manufacturing and Robotics, and the State Key Laboratory of Turbulence and Complex Systems, College of Engineering, Peking University, Beijing, China.}
\thanks{* Send all correspondence to F. Zhang and T. Shen.}
}

\begin{document}


\maketitle

\begin{abstract}
   The artificial lateral line (ALL), comprising distributed flow sensors, has been successful in sensing motion states of bioinspired underwater robots like robotic fish. However, its application to robots driven by rotating propellers remains unexplored due to the complexity of propeller wake flow. This paper investigates the feasibility of using ALL to sense propeller wake for underwater robot leader-follower formation. To estimate the lateral motion states of a leader propeller, this paper designs a multi-output deep learning network that extracts temporal and spatial features from distributed pressure measurements of propeller wake. Extensive experiments are conducted on a designed testbed, the results of which validate the effectiveness of the proposed propeller wake sensing method.
\end{abstract}

\section{Introduction}

In the nature, the lateral line is an essential class of fish's flow-sensing organ that plays a crucial role in their flow-relative behaviors such as obstacle avoidance, rheotaxis, predation, and schooling\cite{Coombs2001AR,Coombs2005Elsevier,Zhai2021JBE}. Taking the inspiration therein, researchers have designed various artificial lateral line systems (ALLs) aiming to enhance flow-sensing capabilities of bioinspired underwater robots\cite{Abdulsadda2011EAPAD,DeVries2015BiB,Taavi2013PRSA,Liu2014Sensors,Zhang2015BiB,Yen2018JOE,Dang2021RAL,Jiang2022Tmech}. 
Most existing research on the ALL has focused on sensing flow fields generated by bioinspired robotic fish with body/fin undulations or dipoles with periodic oscillations. Generally speaking, such flow fields are tractable and represented by either analytical potential flow models\cite{Wang2022CDC,Yen2022BiB} or regular vortex street models\cite{Gao2018JFM,Colvert2018BiB}. 

However, most underwater robots used in the real world use propellers for propulsion \cite{Lowndes2020PHD,Amory2016OCEANS,Rypkema2019MIT}.
The wake of high-speed rotating propellers is complex, dynamic and even stochastic \mbox{\cite{Kumar2017JFM,Felli2011JFM,Wang2021POF}}. Conventional instantaneous pressure measurements are insufficient in providing usable information for flow sensing, while only flow features extracted from the pressure measurement time series are potentially useful. Additionally, analyzing propeller wake demands high-precision and high-frequency sampling \cite{Jie2022OE,Wang2020OE,Long2022ASME}, which is costly and impractical for real-time robot sensing. The distributed low-cost pressure sensors, adopted in ALLs, capture limited dynamics information of the propeller wake at a frequency much lower than the rotating frequency of the propeller. This complexity challenges traditional estimation methods especially the commonly-adopted frequency analysis-based approach \cite{Klein2011BJN,Rodwell2023BiB,Qiu2023TASE,Liu2022SNA}.
Therefore, although many successful designs and applications of ALL have been demonstrated, there remains an open question of whether ALL can provide sufficient feedback information for flow sensing of propeller-driven underwater robots, and if yes, how to design an estimation algorithm to assimilate the distributed measurements.

To address the aforementioned problem, this paper investigates the feasibility and the algorithmic design of sensing the wake flow of a high-speed rotating propeller using an ALL. This paper takes a bold hypothesis that the distributed pressure measurement data sampled by the ALL from the highly dynamic and complex wake flow, although cannot represent the entire flow dynamics, is still sufficiently rich for estimating the relative motion states of the leader propeller-driven underwater robot in a leader-follower formation. While the relative longitudinal or forward-backward movement is an important factor in the leader-follower formation control, the lateral or side-to-side movement is more critical for the success of following the leader using the ALL. Therefore, this paper focuses on the estimation of relative lateral motion states of a leader underwater robot. 

This paper first designs an ALL with low-cost commercial pressure sensors and a corresponding testbed with data collection devices and motion control module. Extensive experiments are conducted using the designed testbed in a testing pool. To address the problem of flow sensing in the highly dynamic and stochastic propeller wake, this paper adopts a one-dimension convolution neutral network (1DCNN) and a bi-directional long short-term memory neural network (BiLSTM) to extract the spatiotemporal characteristics of the time series data of the pressure measurements collected by the ALL. To simultaneously estimate multiple lateral motion states of a propeller-driven leader robot including the displacement, the velocity magnitude, and the direction of the velocity, a multi-output deep learning network is adopted and trained based on the experimental data. The whale optimization algorithm (WOA) is used to optimize the task weights to improve the overall estimation performance. 

The main contributions of this paper are twofold. Firstly, by leveraging an artificial lateral line (ALL) composed of distributed commercial low-cost pressure sensors, this study explores a novel realm of sensing the highly dynamic and complex propeller wake to estimate the position of a leader high-speed rotating propeller. The wake flow under investigation is encountered more commonly by existing underwater vehicles and differs inherently from the flow fields studied in other ALL flow sensing research, such as those generated by bioinspired body undulation and/or tail flapping.
Secondly, this paper proposes a novel data-driven flow sensing algorithm for estimating the lateral motion states of a leader propeller. This algorithm combines deep learning with a biomimetic optimizer to address the complex and dynamic propeller wake sensing.

\section{Problem Description}

\begin{figure}[htp!]
    \centering
    \includegraphics[width=0.99\linewidth]{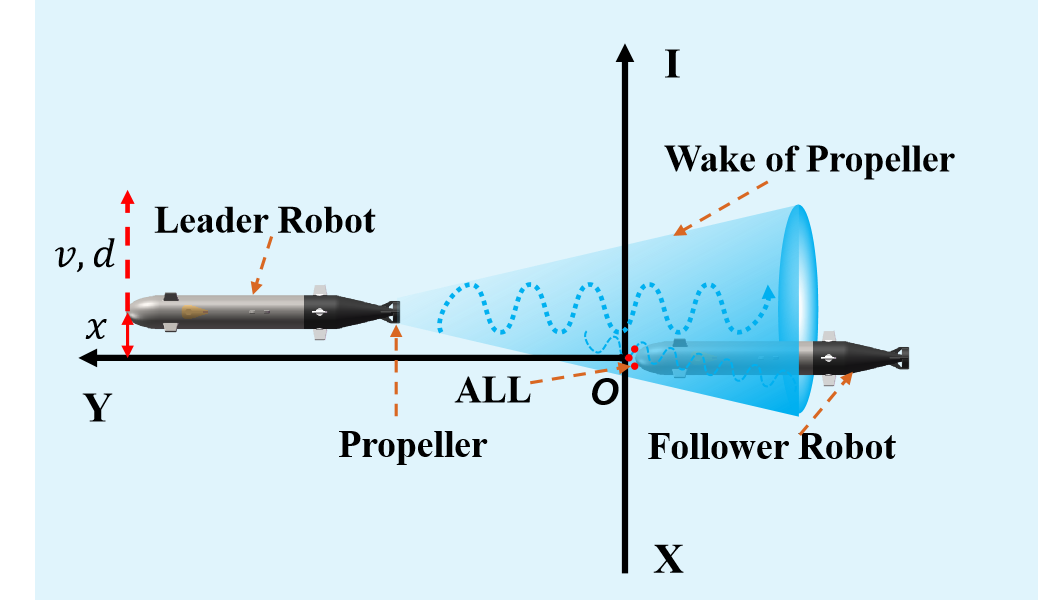}
    \caption{Illustration of ALL sensing the propeller wake of a leader underwater robot in a leader-follower formation (Top View).}
    \label{fig:auv}
\end{figure}

Leader-follower formation is one of the most commonly-used formation configuration in underwater robots\cite{Jia2015Tmech}. Analogous to the lateral line of biological fish in sensing a leader fish, the ALL is expected to estimate the corresponding motion states of the leader robot by sampling its wake flow (Fig.~\ref{fig:auv}). Particularly, we are more interested in sensing the lateral motion of the leader compared to the longitudinal motion which has a significant influence on the formation control. 
In addition, this paper replaces the leader robot with a single propeller with an aim to focus on the fundamental research problem in propeller wake sensing. While we acknowledge that wake flows generated by traveling robot bodies affect those produced by propellers, many existing studies\mbox{\cite{Zheng2017BiB,Rodwell2023BiB,Free2018BiB}} on ALL-based motion state estimation adopted the same simplification measures. The underlying reason is that wake flows generated by propulsion systems typically possess significantly different features than flow fields generated by traveling robot bodies, therefore, are detectable when mixed with vortices shed by traveling robots.
In this paper, we hypothesize that the spatiotemporal flow measurement data acquired by the ALL, although only contains partial coherent flow information of the highly dynamic propeller wake, holds sufficient information regarding the lateral motion of the leader robot, which is described by the velocity magnitude, the velocity direction and the displacement.

\subsection{Testing Platform}
\begin{figure*}[!htb]
    \centering
    \subfigure[CAD design of the testing platform.]
    {\includegraphics[width=0.55\hsize]{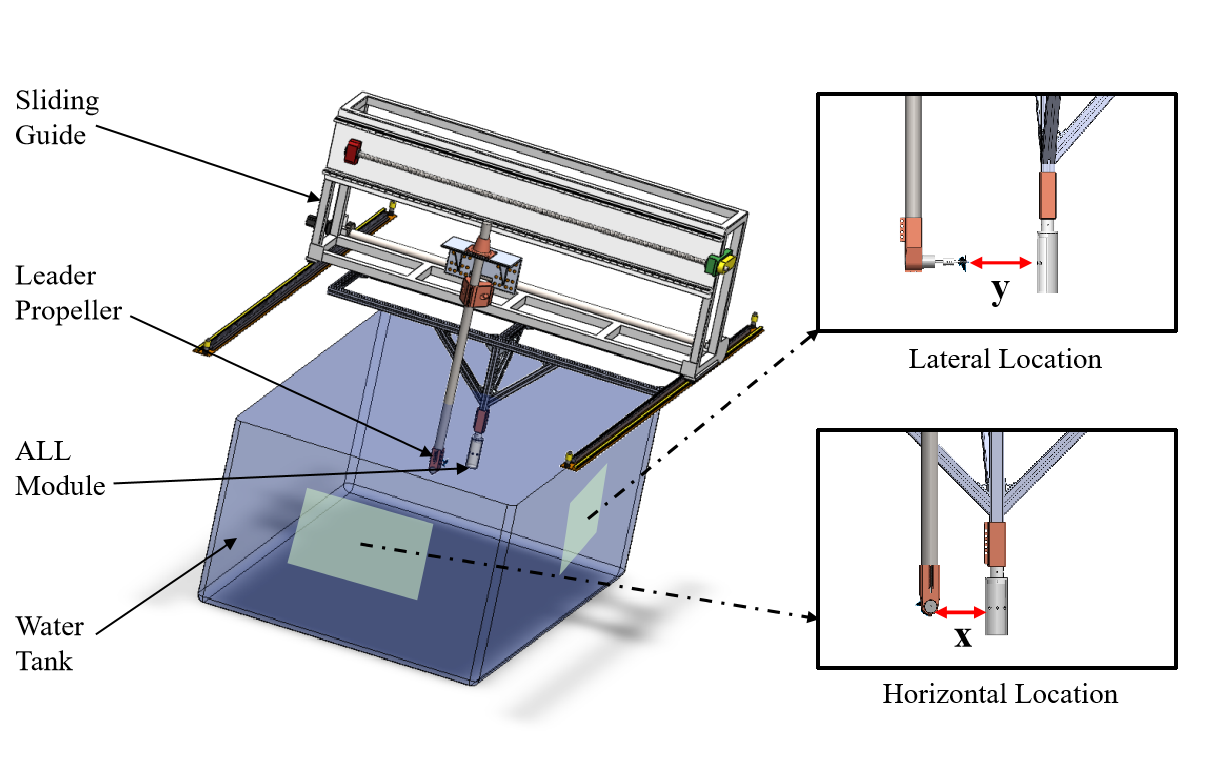}\label{fig:testbad}}
    \subfigure[ALL design with pressure sensors.]{\includegraphics[width=0.4\hsize]{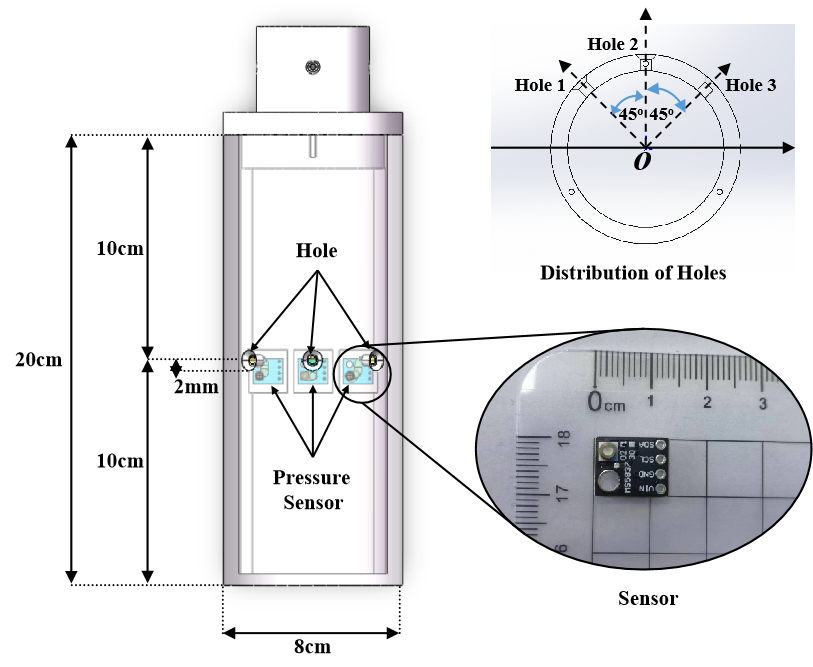}\label{fig:ALL}}
    \caption{The design of the testing platform that comprises a water tank, a sliding guide, a leader propeller, and an ALL. The water tank measures 3.0m(L) x 2.0m(W) x 1.5m(H). The ALL consists of three pressure sensors distributively positioned with an angle of $45^\circ$ between each adjacent pair at the same horizontal level from the bottom of the cylindrical shell.}
    \label{fig:scenario}
\end{figure*}

The wake of a propeller-driven underwater robot traveling through aquatic environments is mainly dominated by the propeller wake, therefore, this paper simplifies the problem and focuses on the estimation of the motion states of a leading propeller. For this purpose, we have constructed a testing apparatus shown in Fig.~\ref{fig:scenario}. This experimental testbed mainly consists of a testing water tank, a high-precision sliding guide, a data acquisition system and a leader propeller. The water tank is 3m long, 2m wide and 1.5m deep. The sliding guide provides support and high-precision movement between the rotating propeller and the ALL system. A high-performance industrial personal computer (IPC) controls the sliding guide with a linear traveling velocity between 0.2~m/min and 1.2~m/min and a positional accuracy of 1~mm. Similarly to a towing tank, the sliding guide is rigidly attached to the water tank and is controllable through preprogramming or manual operation.

Attached to the sliding guide are a data acquisition system and a leader propeller system. The data acquisition system comprises an ALL equipped with three pressure sensors and a microcomputer. A circular shaped cross-section is widely used in ALL flow sensing for its corresponding tractable hydrodynamic properties. This paper adopts the same circular shape design for the ALL (as shown in Fig.~\ref{fig:ALL}). The cylindrical ALL has a diameter of 8 cm and a height of 20 cm. Three through-holes with a radius of 2 mm are evenly placed for holding pressure sensors with an angle of 45 degrees between each adjacent pair at the same horizontal level of 10 cm from the bottom of the cylinder. The pressure sensors used are MS5837-02Ba,  featuring a resolution of 1 Pa and I2C bus communication. To acquire multiple sensors simultaneously in real time, an TI TCA9548 eight-channel bidirectional transfer switch is used. Moreover, the two subfigures of Fig.~\ref{fig:testbad} show the relative position relationship between ALL and the propeller.

To accurately measure the positional states of the leading propeller, we installed a high-precision laser ranging sensor (LGKG Co HG-C1400-P). This ranging sensor features a measurement range of 0-400 mm and an accuracy of 0.8~mm with an output frequency of 100 Hz. The outputs of the ranging sensors and the ALL pressure sensors are acquired by an Arduino micro-controller that synchronizes all the signals and sends all the real-time data to the host computer, a Raspberry Pi 4 Model B with 8G memory that operates in Linux (Ubuntu 18.04) a Python program for data acquisition, processing and storage.

\subsection{Estimation Problem}

Taking the assumption that the rotating axis of the leader propeller and all the pressure sensors of the ALL locate in the same horizontal plane, we project the propeller and the ALL onto that horizontal plane. The ALL is represented by a circle under the projection and the three sensors are represented by three points on the circle. We define a reference frame $I$ within the horizontal plane, the center of the circle as the origin $O$, and the axis pointing from the origin to the center pressure sensor as the $y$-axis. The $x$-axis is perpendicular to the $y$-axis pointing to the left. 
In experiment, the rotating axis of the propeller is kept parallel to the $y$-axis. The displacement of the propeller with respect to the origin is denoted by $(x,y)$ where $x$ and $y$ represent the lateral and longitudinal displacements of the leader propeller, respectively. 
We denote the speed of the propeller's lateral motion along the $x$-axis by $v$, and the corresponding direction of the motion by the boollean-type varaible $d=\{\mathrm{P\ or\ N\}}$ indicating the positive or negative direction along the $x$-axis. 
The lateral motion states of the propeller, denoted by $\boldsymbol{s}$, include three variables, i.e., $\boldsymbol{s} =\left [ x, v,d \right ]^{\mathrm{T}}$. 
The measurement of the ALL distributed pressure sensors at time $k$ is defined as
$\boldsymbol{m}_k\in \mathbb{R}^{N}$
where $N$ represents the number of the ALL pressure sensors. 
With the distributed pressure sensors measuring their corresponding local pressures in the propeller wake, the objective is to design a flow-sensing algorithm represented by function $f$ such that
\begin{equation}
    \lim_{k \to \infty} \left | \hat{\boldsymbol{s}}_k - \boldsymbol{s}_k \right | =0
\end{equation}
where $\hat{\boldsymbol{s}}_k =f(\boldsymbol{m}_k)$  calculates the estimated motion states $\hat{\boldsymbol{s}}$ using the ALL pressure measurements $\boldsymbol{m}_k$.

\section{Dataset Construction}
\subsection{Data Acquisition}

We conduct extensive experiments using the aforementioned testbed to create a dataset of the ALL measurements given selected lateral motion states of the leader propeller. To eliminate the influence of the submerged depth of the pressure sensors, all the sensor outputs are debiased by deducting the corresponding pressure readings sampled before the rotation of the propeller when the flow is stationary. In each experimental trial, the leader propeller moves along the lateral axis or the $x$-axis at a constant speed $v$ with motion direction $d\in\{N,P\}$ while the longitudinal displacement $y$ is kept constant. 
We select five representative speeds of the moving propeller, i.e., $v\in\{400, 500, 600, 700, 800\}$ mm/s. 
Considering the effective measurable width of the propeller wake by the ALL and the influence of the transient during the propeller accelerating period,  the lateral motion of the propeller goes from $x=-175 $~mm to $x=175 $~mm and then back to $x=-175 $~mm. The effective estimation range is clipped to $x\in[-120, 120]$~mm and normalized in the following sections for the convenience of presentation. 
In addition, two longitudinal displacements are selected in the experiment, i.e., $y \in \{250, 300\}$~mm.

The back-and-forth trip is repeated 10 times for each set of motion state configurations.
\begin{figure}[!htb]
    \centering
    \subfigure[Positive direction of motion or $d=P$.]{\includegraphics[width=\hsize]{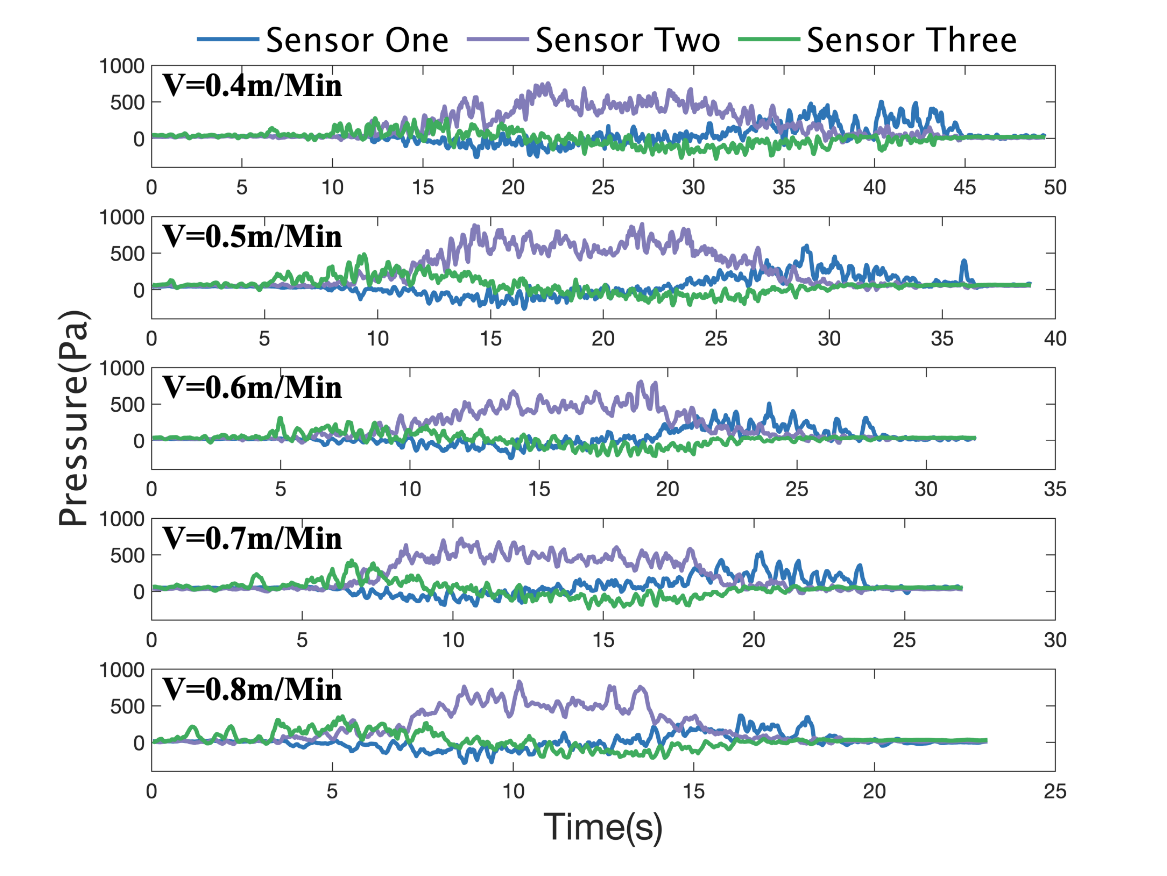}\label{fig: dyflcu2}}
    \subfigure[Negative direction of motion or $d=N$.]{\includegraphics[width=\hsize]{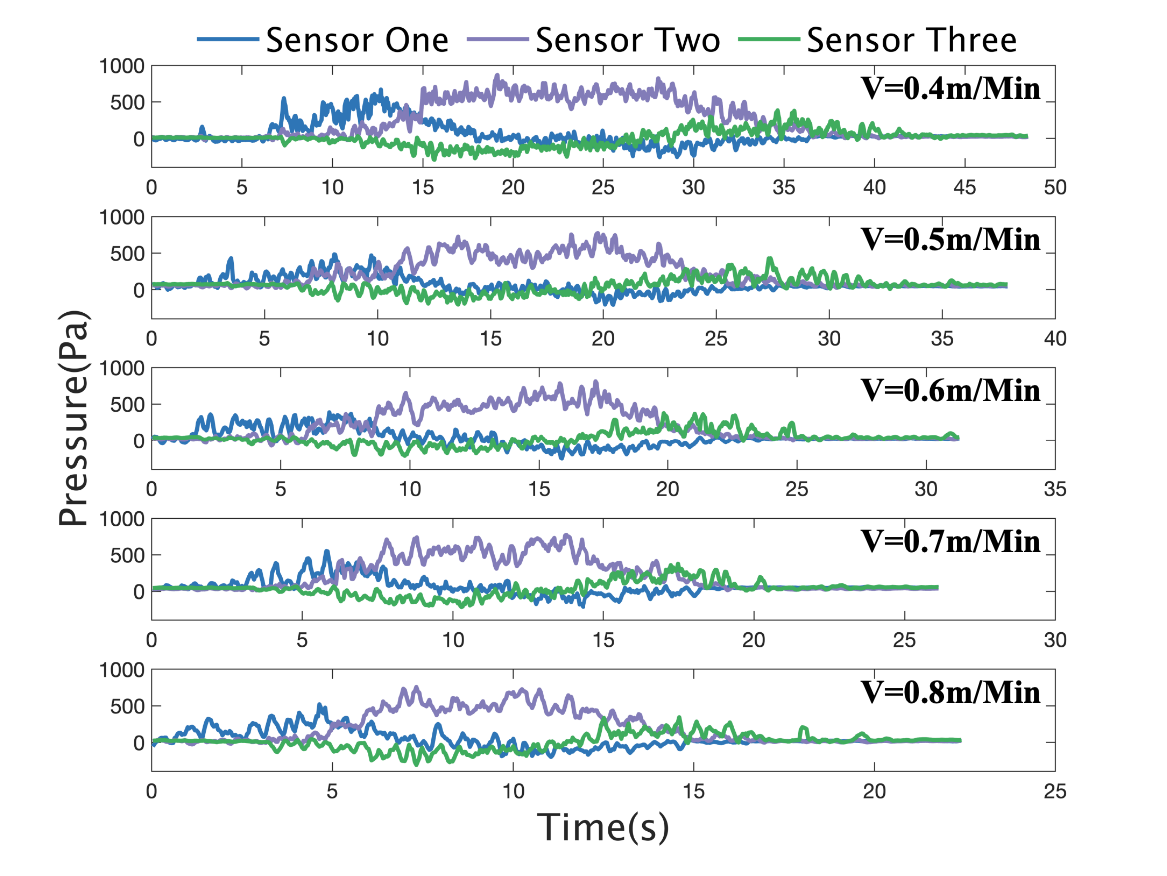}\label{fig: dyflcu4}}
    \caption{The trajectories of pressure sensor data collected by the ALL in the experiment when the leader propeller moves laterally at different speeds.}
    \label{fig: dynamicflcut}
\end{figure}
Figure~\ref{fig: dynamicflcut} shows the collected distributed pressure data given selected speeds and directions of the leader propeller. 
We observe strong correlations between time-domain features of the pressure measurements and the lateral motion states of interest. Specifically, the changing trends of the collective pressure measurements are primarily correlated to the direction of motion of the leader propeller; the changing rates of the pressure measurements are mainly correlated to the traveling speed of the propeller; and the magnitudes of the pressure measurements are mainly correlated to the lateral displacement of the propeller. Such observations indicate that the ALL pressure measurements of the propeller wake (most likely) contain sufficiently rich information for the estimation of the lateral motion states of the leader propeller, supporting our previous hypothesis.

\subsection{Time Series Selection}
Most existing flow sensing studies using ALLs deal with structured flow fields such as the Von Karman vortex street of laminar flow passing a cylinder or periodic wake flow of an undulating fish-like fin. This paper, however, targets the propeller wake which features cross-scale dynamic and unstable flow structures, thus making the flow sensing problem extremely challenging. 
Due to the practical limitation of sampling frequency and sampling resolution of the ALL, it is infeasible to accurately capture all the characteristics of the propeller wake. 
Additionally, the phase averaging approach commonly adopted in propeller wake analysis requires the data sampling to occur at uniform phases of the periodic rotation of the propeller, which exceeds the sampling capability of almost all the ALL sensing systems.
Through high-fidelity CFD simulation, we find that during the motion of the propeller, there exists a stochastic pulsation process in the ALL pressure output along with statistically distinguishable trends corresponding to different motion configurations. 
Therefore, we hypothesize that the time series of ALL measurement data provides partial but sufficient information regarding the lateral motion of the propeller of interest. Furthermore, increasing the number of sensors and/or the total sampling time leads to added information acquired, enhancing the robustness of the flow sensing algorithm, which poses a balancing problem between the practical engineering considerations (e.g., the cost, the installation space and real-time calculation) and the estimation accuracy.
Specifically, we define the time series of the $i$-th pressure sensor output at time $k$ used for wake flow sensing as 
\begin{equation}
    \boldsymbol{M}_{k} = \left [ \boldsymbol{m}_{k-sl+1},...,\boldsymbol{m}_k \right ]^\mathrm{T}
\end{equation}
where $\boldsymbol{m}_k\in \mathbb{R}^{N}$ represents the distributed pressure readings at time $k$, and $sl$ represents the sequence length of the time series. The sequential data $\boldsymbol{M}_k$ continuously evolves with upcoming sensory measurements embedding the flow dynamics information of the latest $sl$ time steps, forming the dataset of the time series of the ALL distributed pressure measurements for algorithm training and testing.

\begin{figure*}[ht!]
    \centering    \includegraphics[width=0.85\textwidth]{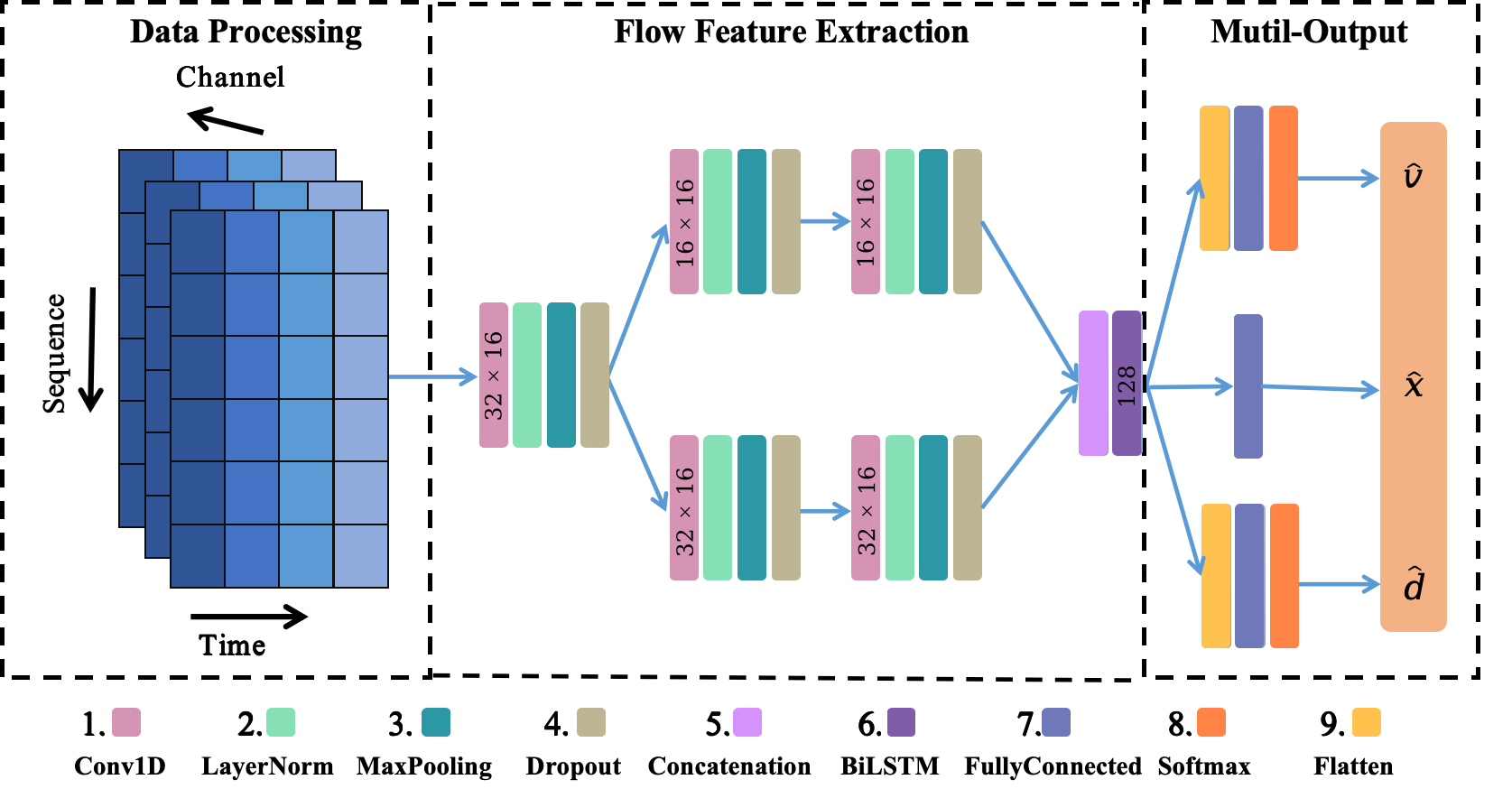}
    \caption{The schematic of the network architecture of the proposed wake flow estimator that consists of a hybrid CNN-BiLSTM network for flow feature extraction and a multi-output neural network for motion state regression and classification.}
    \label{fig:newwork}
\end{figure*}
\section{Motion State Estimation via Wake Sensing}
\subsection{Flow Feature Extraction with Hybrid CNN-BiLSTM}
Considering the highly complex and stochastic dynamics of propeller wake, we adopt the trending powerful deep learning approach to extract hidden spatiotemporal flow features given the time series of the distributed ALL measurements and design a hybrid function of the one-dimension convolutional neural network (1DCNN) and the bidirectional long short-term memory network (BiLSTM).
Specifically, as one of the most widely used networks\cite{Kiranyaz2021MSSP,Jayanth2016arxiv}, 1DCNN is selected to extract hidden spatial features of the sequential pressure measurement data supported by sparse connections and weight sharing. The network design consists of four main layers including the input, the convolutional, the batch normalization, and the pooling (Fig.~\ref{fig:newwork}).
On the other hand, BiLSTM is a special recurrent neural network (RNN)\cite{Hochreiter1997NC,Graves2005NN} which introduces a gate mechanism to control both the forward and reverse information transmission paths, rendering BiLSTM a stronger learning ability for long-range dependency compared to traditional RNNs\cite{Hochreiter1997NC}. In this paper, the BiLSTM network follows 1DCNN and learns the bidirectional sequential relationship from the extracted spatial features, thereby providing long-term characteristics of the time series of ALL data sampled from the unsteady and complex wake flow of the leader propeller.

Figure~\ref{fig:newwork} shows the network architecture of the designed hybrid CNN-BiLSTM model. Firstly, the time series of distributed sensor data is input into a multi-channel 1DCNN where feature extraction and dimension reduction are conducted through the convolution and pooling operations. Next, the output of the 1DCNN network is fed into the BiLSTM network\cite{Li2024SP,Mohine2022TITS} whose forgetting, input, and output gates are optimized via iterative training to learn the time-domain relationship between spatial features extracted by the 1DCNN network. Therefore, the hybrid 1DCNN-BiLSTM network extracts spatiotemporal features of the sampled wake data, expectedly providing sufficient information for estimating the lateral motion states of the leader propeller.

\subsection{Motion State Estimator}
\textbf{Design of the loss function.}
This paper focuses on the estimation of the lateral motion states of the leader propeller including the relative displacement $x$, traveling speed $v$, and direction of motion $d$. Considering the properties of the motion states, we model the estimation task as a combination of regression and classification problems with continuous estimation state space of displacement $\mathcal{X}$ and discretized estimation state space of traveling speed $\mathcal{V}$ and Boolean-type direction $\mathcal{D}$. Correspondingly, we design a multi-output neural network as the motion state estimator that takes the wake flow features as input and estimates the lateral motion states of the leader propeller utilizing parameter sharing \cite{Bao2023Energies}. The network architecture of the designed multi-output model is depicted in the right column of Fig.~\ref{fig:newwork}. Given sufficiently rich experimental data, the key to the design of the state estimation model lies in the design of the loss function \cite{Li2022ECTI}. Considering the continuous and discrete motion states to be estimated, we first design three individual loss functions for the three estimation states, respectively. Specifically, the relative lateral displacement $x$ is estimated by regression, with its loss function $l_1$ designed as the mean square error loss (MSE), i.e.,
\begin{equation}
    l_1=\frac{1}{N} \sum_{i=1}^{N}\left(\hat{x} -x\right)^{2}
\end{equation}
where $x$ and $\hat{x}$ represent the actual and the estimated displacements, respectively.
Modeled as discrete states, the speed $v$ and the direction of speed $d$ are estimated via classification with their loss functions $l_2$ and $l_3$ designed using the cross-entropy loss, i.e.,
\begin{equation}
    l_2=-\frac{1}{N} \sum_{i}^{N} \sum_{k=1}^{V} z_{i k} \log \left(p_{i k}\right)\\ 
\end{equation}
\begin{equation}
    l_3=-\frac{1}{N} \sum_{i}^{N} \sum_{k=1}^{D} z_{i k} \log \left(p_{i k}\right)
\end{equation}
where $N$ is the size of the sampling batch, $V$ and $D$ are the numbers of categories or the discretization levels of the speed and direction of motion, respectively, boolean-type $z_{i k}$ denote whether sample $i$ belongs to category $k$ (1 for yes and 0 for no), $p_{i k}$ denotes the predicted probability that the observed sample $i$ belongs to category $k$.

Designed as a linear combination of the three individual losses, the overall loss function of the state estimator calculates as follows
\begin{equation}
    l=\lambda_{1} l_{1}+\lambda_{2} l_{2}+\lambda_{3} l_{3}
\end{equation}
where the task weight $\lambda_i$ balances the relative importance between the three estimated states. The multiple output model utilizes the feature extraction module with parameter sharing.

\textbf{Optimizing task weights.}
To ensure a balanced influence of the task weights on the estimation performance, we formulate weight selection as a multi-objective optimization problem. Considering the network complexity and extended computational time, we choose the Whale Optimization Algorithm (WOA) due to its effectiveness in quickly finding optimal solutions. Inspired by humpback whales' predatory behaviors, WOA offers advantages like minimal adjustable parameters, straightforward operation, and a robust ability to escape local optima \cite{Seyedali2016AEF,Lakshmi2021ASC,Nadimi2023ACME,Fu2022SR,Shao2022ABB,Fang2024JES}.
To optimize the task weights, we design a performance evaluation vector $\boldsymbol{e}=[1-\text{ACC}_1,1-\text{ACC}_2,\text{ERR}]$ where $\text{ACC}_1$ and $\text{ACC}_2$ represent the state estimation accuracy of the lateral motion speed $v$ and the direction of motion $d$, respectively, and $\text{ERR}$ represents the mean square error of the estimated displacement $x$. The optimization fitness $g$ (or the cost) is then designed as 1-norm of the performance evaluation vector $\boldsymbol{e}$, i.e.,
\begin{equation}
g(\lambda_1,\lambda_2,\lambda_3,)=\|\boldsymbol{e}\|_1
\end{equation}
Searching for the minimum of the fitness over the task weight space, WOA optimizes the overall estimation performances.
Specifically, WOA first initializes the task weights $\lambda_i$ and the WOA algorithm parameters such as the searching agent population and the maximum iteration number. 
Next, given the wake flow dataset constructed in Section~III, WOA computes the fitness $g$ for each searching agent and updates the position of each agent accordingly. Repeat the step until conditions are met such as parametric convergence obtained or maximum iteration steps reached.

\subsection{Hyperparameters and Training Environment}
Hyperparameters are the tunable training parameters that directly affect the learning results of the estimation network model. In this paper, the batch size is set to 64, the learning rate is set to $1\times10^4$, and the number of epochs is set to 200. Adam's algorithm is used for optimization. 
Adopting softmax output, the classification task categorizes the direction of motion into two classes and the lateral speed of the propeller into five classes. The regression task adopts a fully connected network output normalized to the interval [-1,1]. To avoid overfitting, the dropout technique is utilized. The model training and testing are conducted on a desktop computer with the configuration of an Intel 12900K CPU, an Nvidia RTX 3080 Ti GPU, and a 64GB RAM.

\begin{figure}[!htb]
    \centering
    \subfigure[Lateral displacement $\hat{x}$ in Case 1.]{\includegraphics[width=0.49\linewidth]{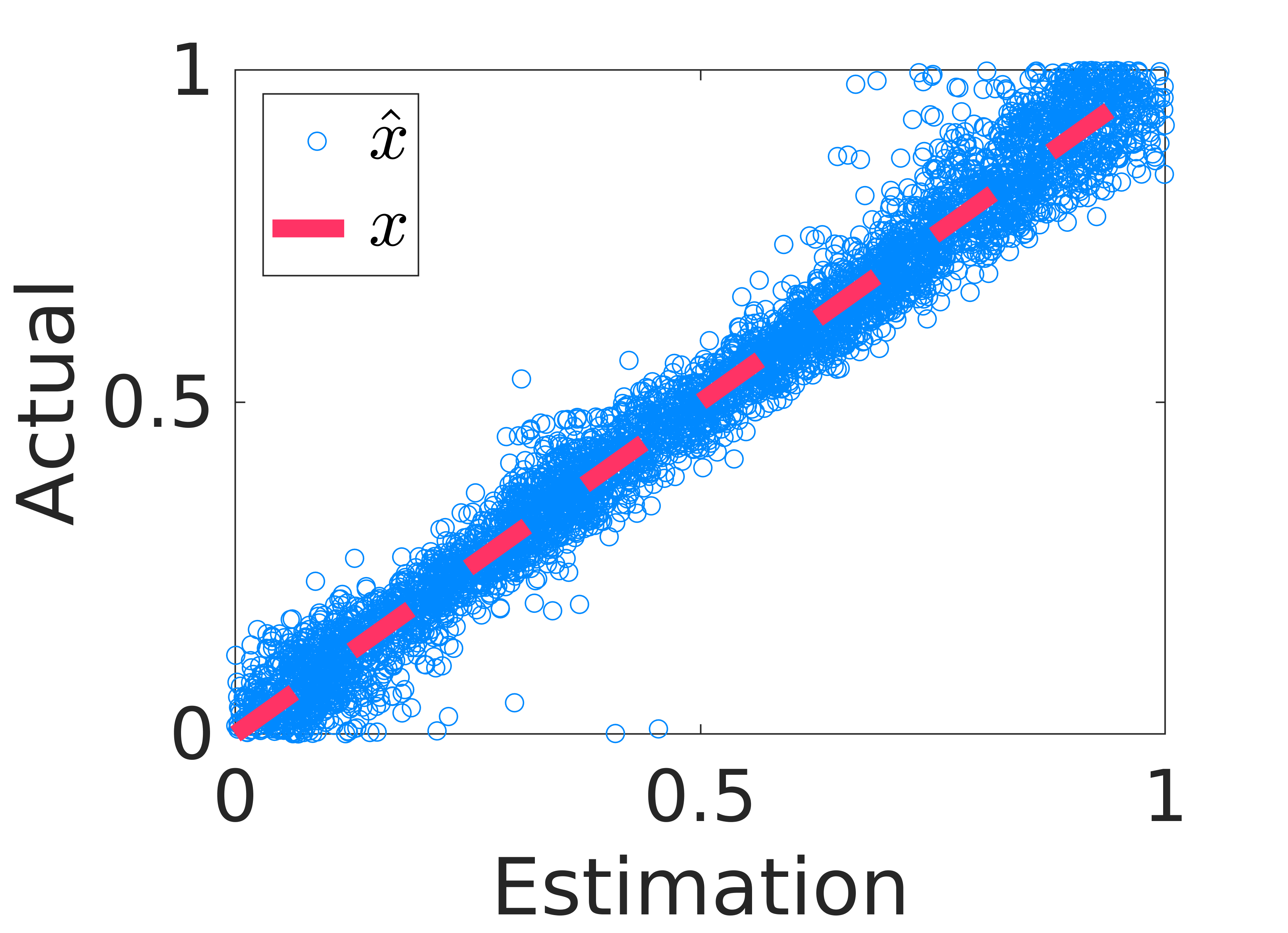}\label{fig: sub_figure1}}
    \subfigure[Lateral displacement $\hat{x}$ in Case 2.]{\includegraphics[width=0.49\linewidth]{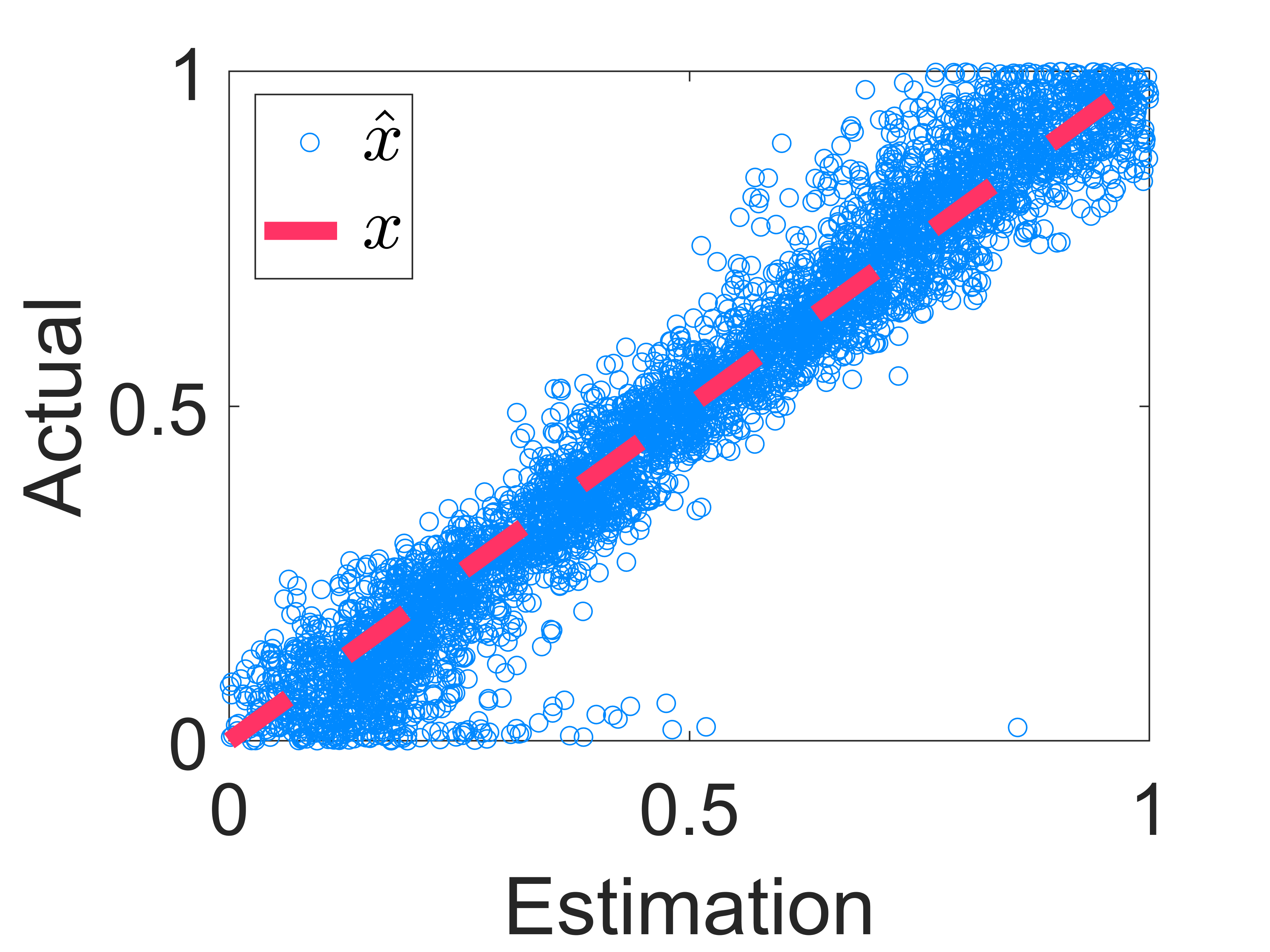}\label{fig: sub_figure1}}
   \subfigure[Confusion matrix of direction of motion $\hat{d}$ in Case 1.]{\includegraphics[width=0.49\linewidth]{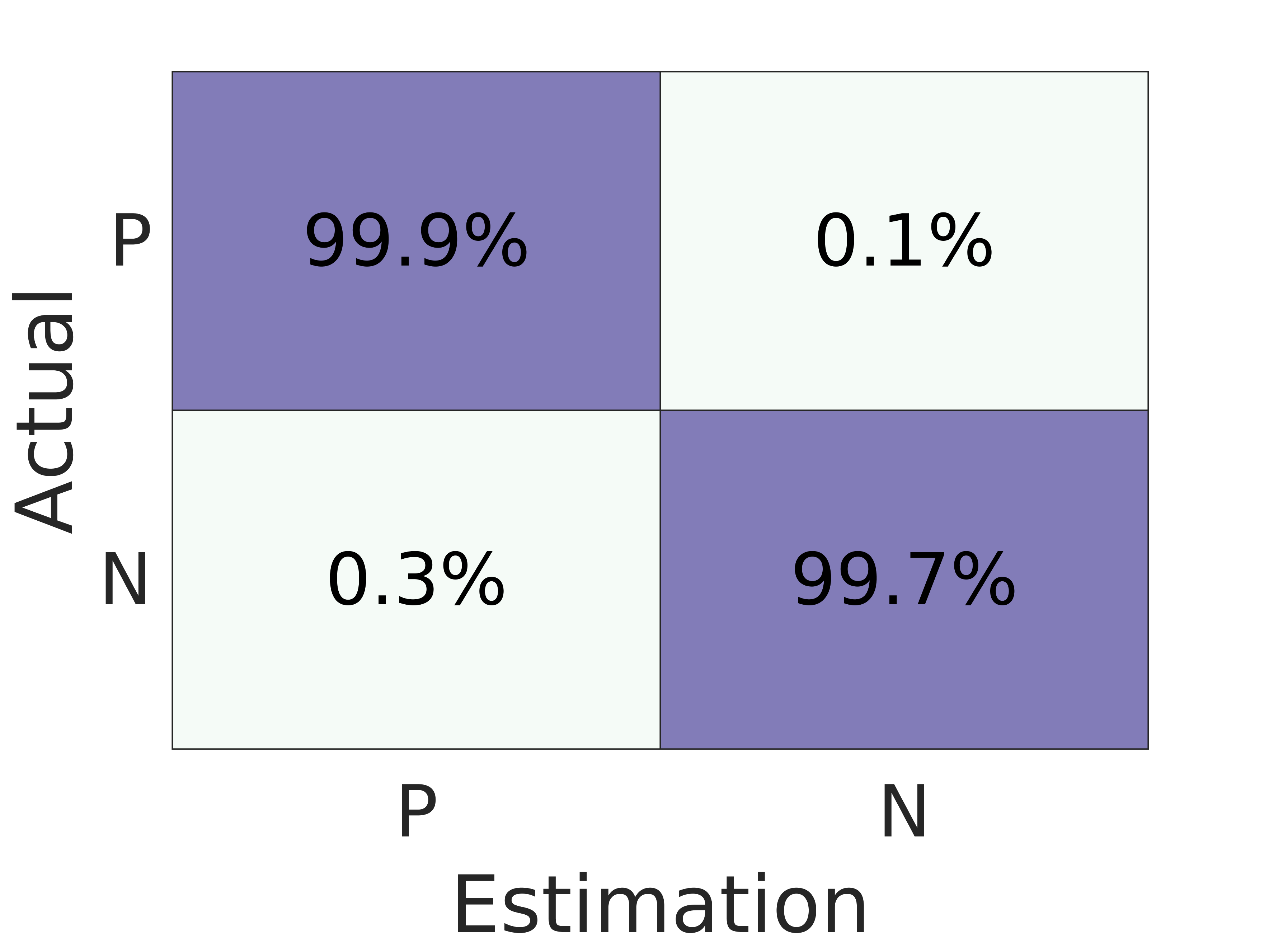}\label{fig: sub_figure3}}
    \subfigure[Confusion matrix of direction of motion $\hat{d}$ in Case 2.]{\includegraphics[width=0.49\linewidth]{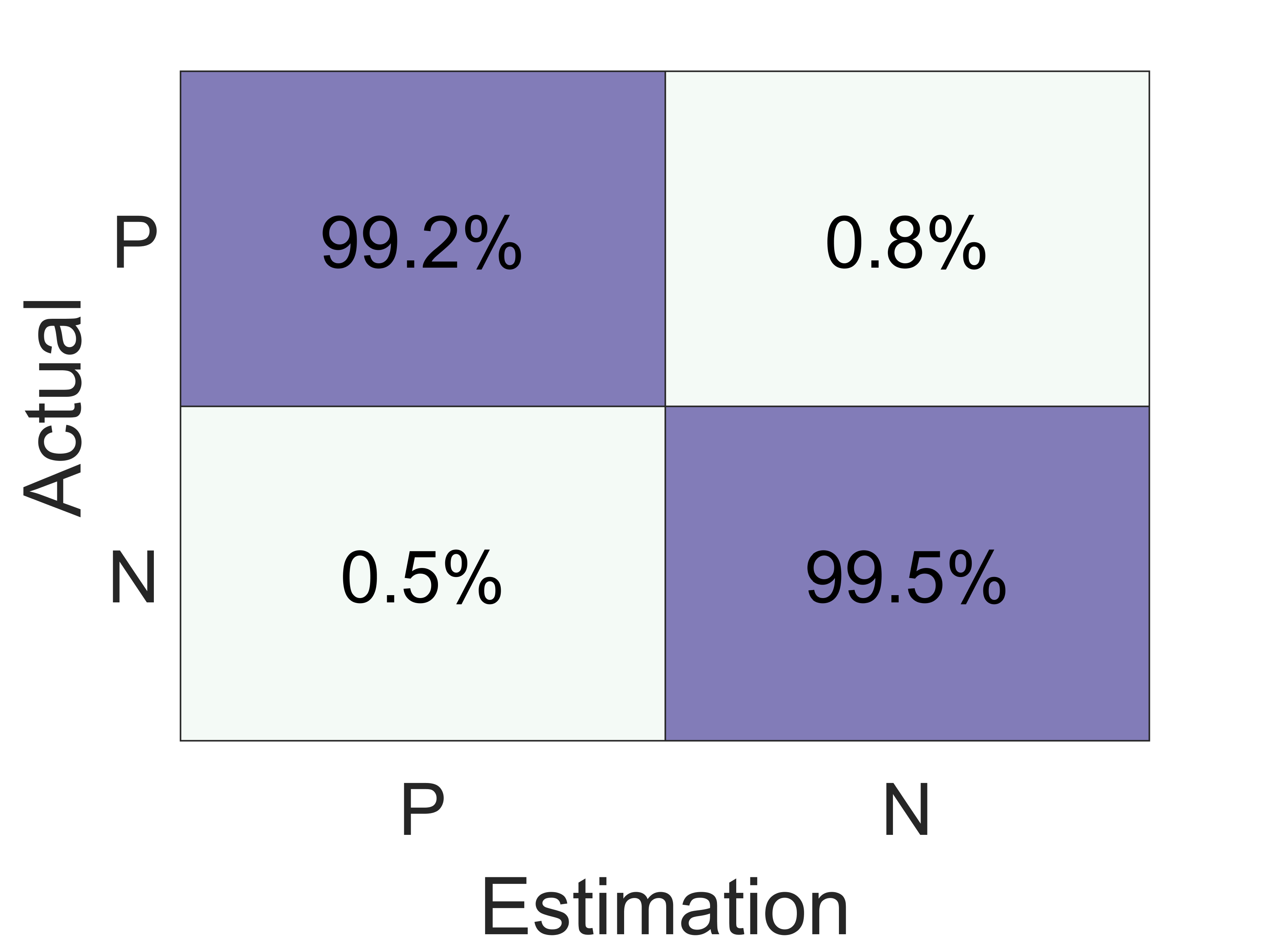}\label{fig: sub_figure3}}
     \subfigure[Confusion matrix of traveling speed $\hat{v}$ in Case 1.]{\includegraphics[width=0.85\linewidth]{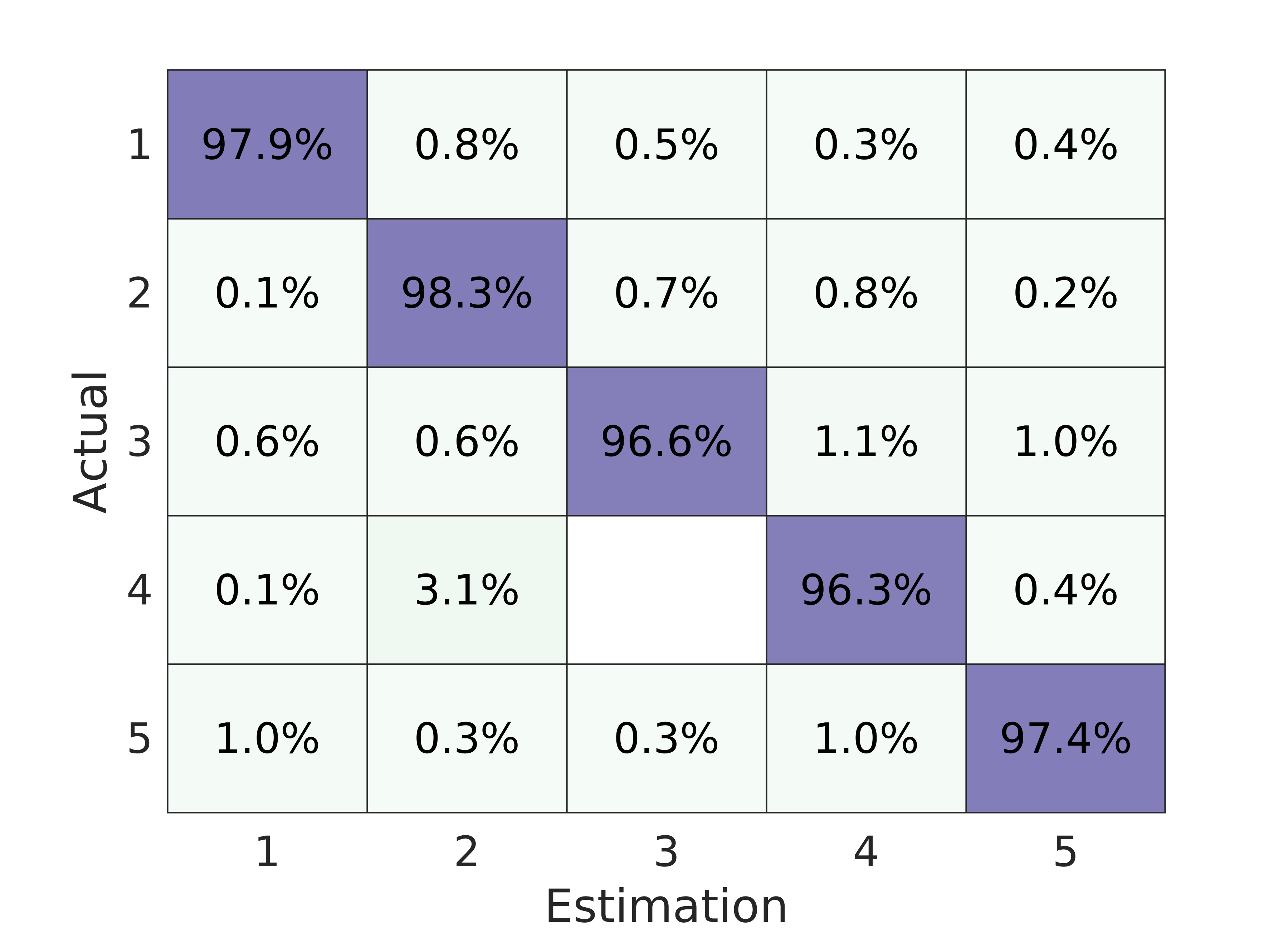}\label{fig: sub_figure2}}
    \subfigure[Confusion matrix of traveling speed $\hat{v}$ in Case 2.]{\includegraphics[width=0.85\linewidth]{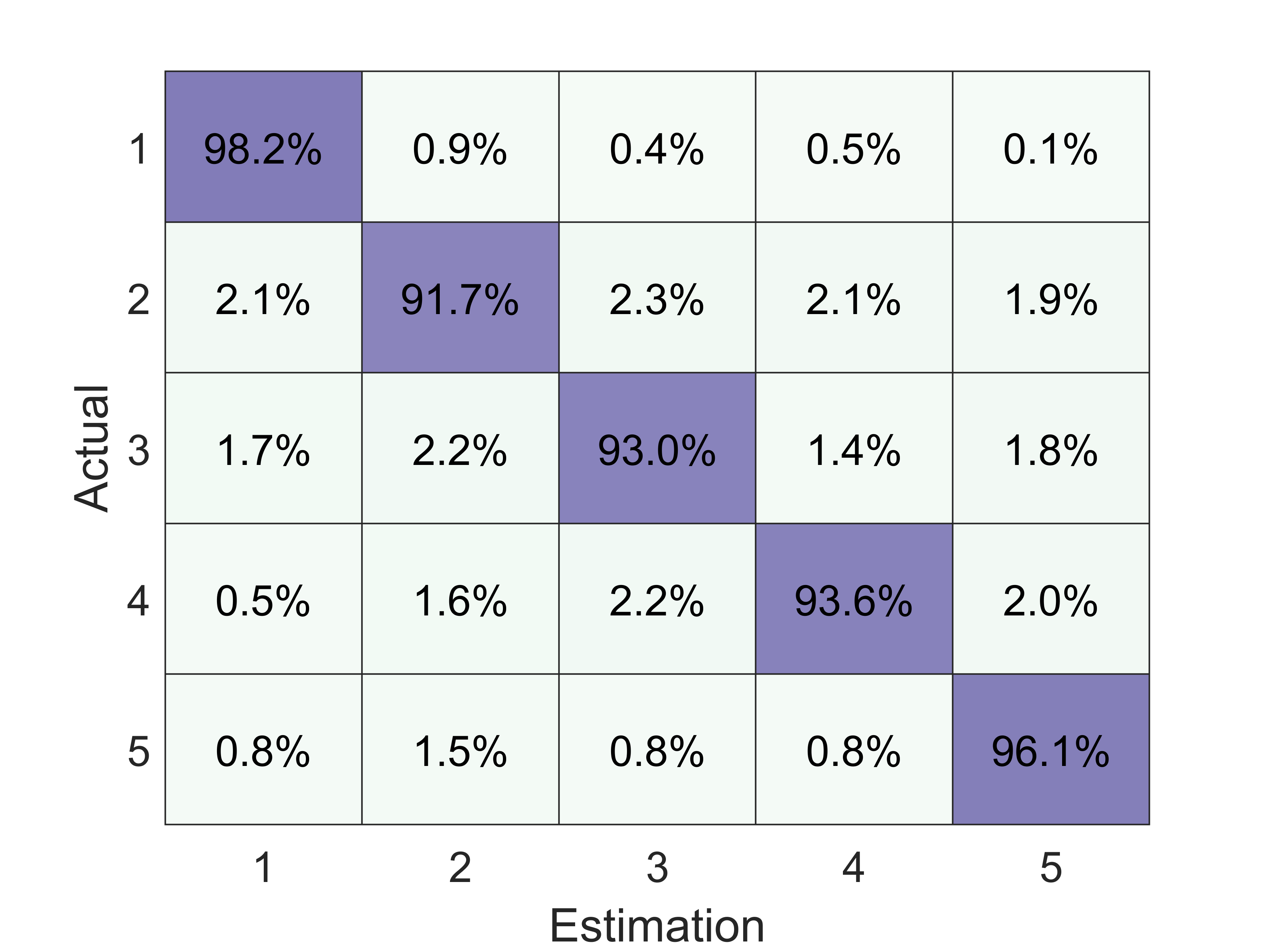}\label{fig: sub_figure2}}
    \caption{The experimental results of the motion state estimation based on propeller wake sensing including the regression results on the lateral displacement $x$, and the classification results on the traveling speed $v$ and the direction of motion $d$.}
    \label{fig: result_est}
\end{figure}

\subsection{Experimental Results}
\begin{figure*}[!htb]
    \centering
    \subfigure[Lateral displacement estimate $\hat{x}$.]{\includegraphics[width=0.32\hsize]{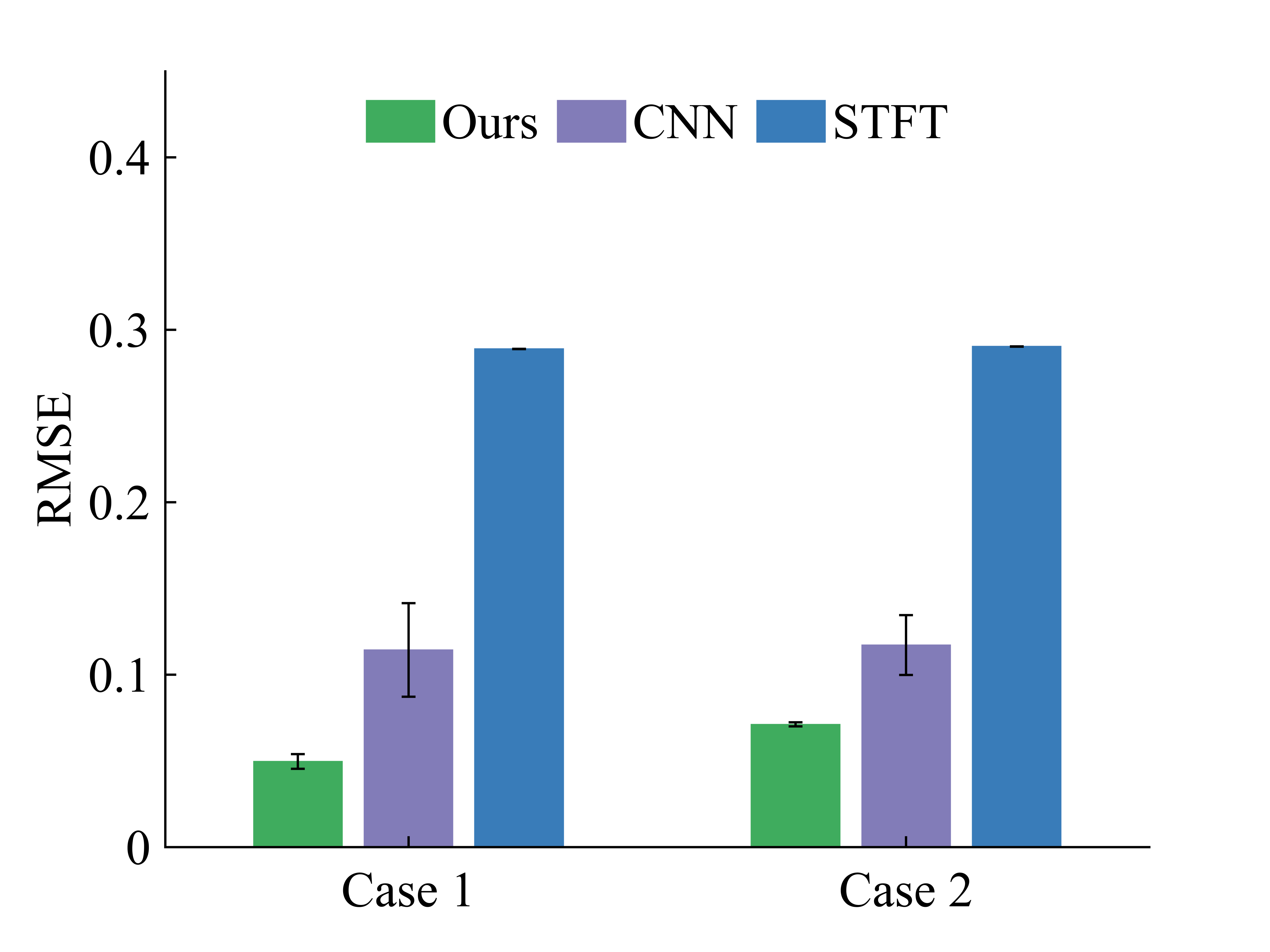}\label{fig: C_RMS}}
    \subfigure[Traveling speed estimate $\hat{v}$.]{\includegraphics[width=0.32\hsize]{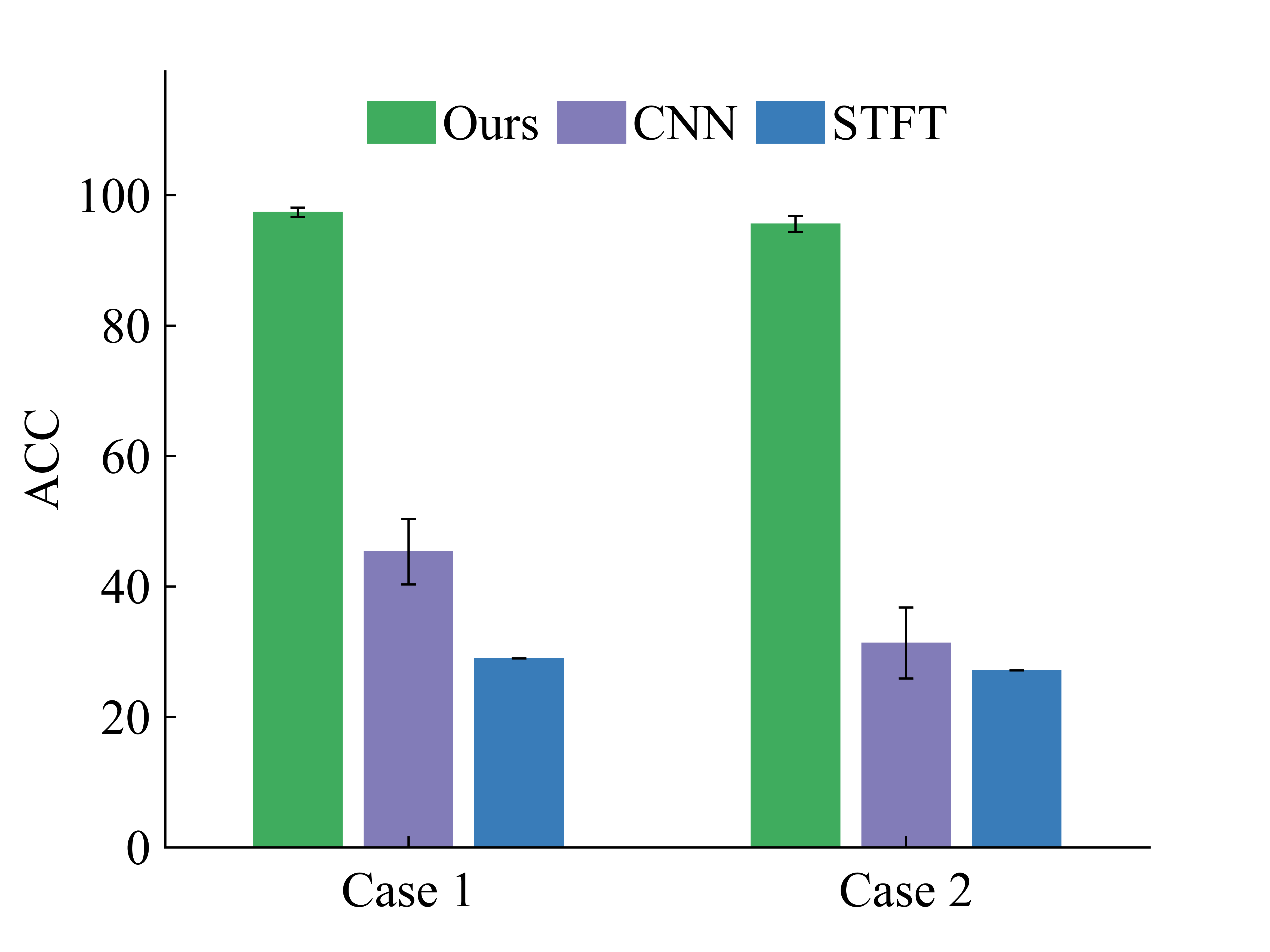}\label{fig:C_ACC2}}
    \subfigure[Direction of motion estimate $\hat{d}$.]{\includegraphics[width=0.32\hsize]{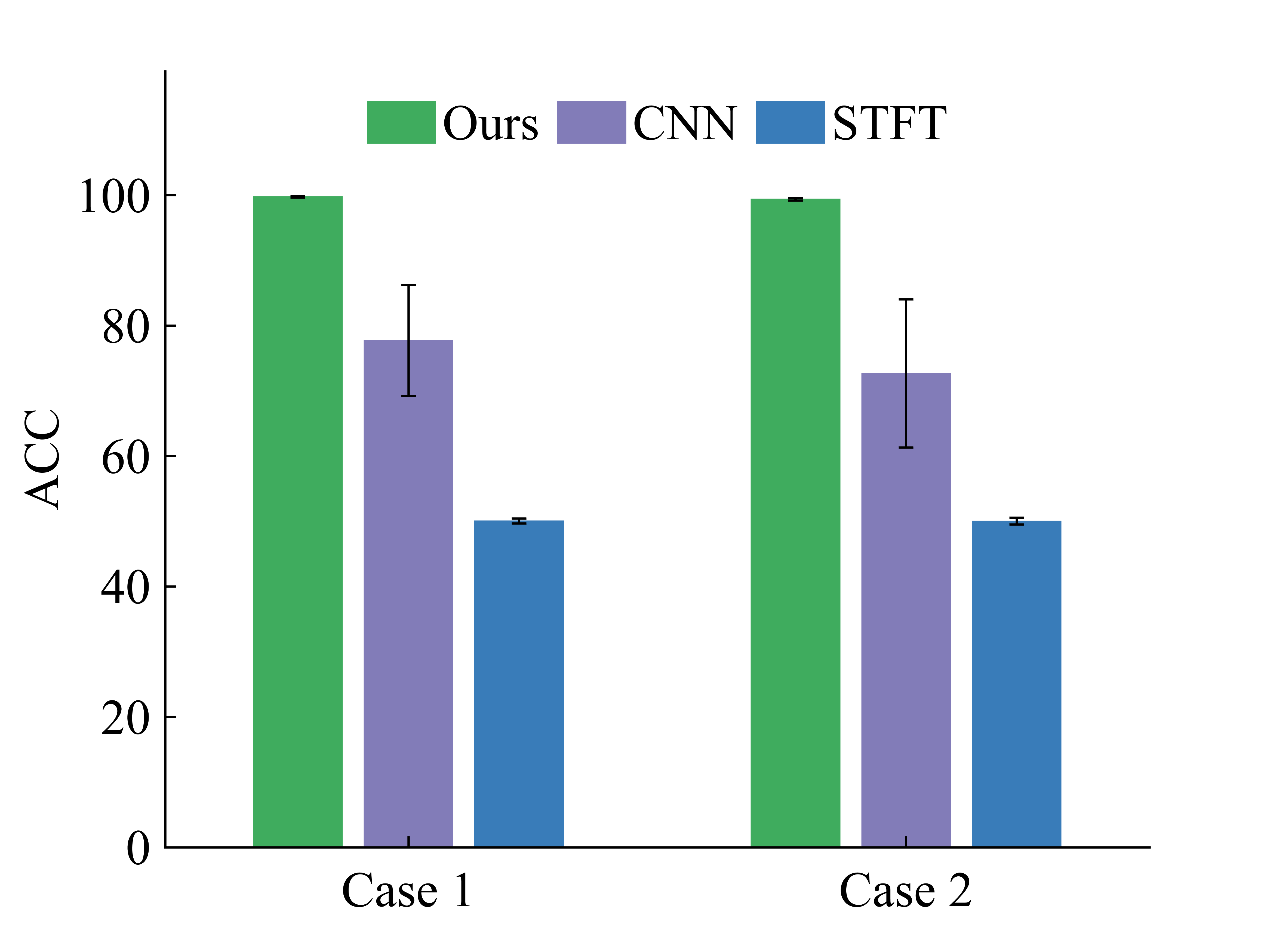}\label{fig: C_ACC1}}
    \caption{The experimental results of motion state estimation comparing the proposed method with traditional frequency analysis-based (STFT) and learning-basded (CNN) methods. Evaluated by the estimation accuracy in terms of all the three motion states of interest, the proposed method consistently outperforms the conventional methods from the literature in addressing the unique propeller wake estimation problem.}
    \label{fig: compare}
\end{figure*}
To demonstrate the effectiveness of the proposed motion state estimation method, experiments are designed and conducted. Two longitudinal displacements $y\in \{250, 300\}$ mm are adopted and tested for each experimental configuration as comparative studies pertaining to the separation distance between the ALL sensing system and the leader propeller. The two longitudinal displacements of $250$~mm and $300$~mm are referred to as Case 1 and Case 2 in the following sections for the convenience of discussion. The length of the sliding window $sl$ is set to $64$. The training and testing data follow a 9:1 ratio division.

Figure \ref{fig: result_est} shows the experimental results of the lateral motion state estimation of the leader propeller. Table~\ref{wake_est_result} presents the statistical estimation results for each motion state over all the experimental tests in Case 1 and Case 2, respectively. Specifically, Figs.~\ref{fig: result_est}(a) and \ref{fig: result_est}(b) show the estimates and the actual values of the lateral displacement of the traveling leader propeller for Case 1 and Case 2, respectively. The red dashed line represents perfect estimation for reference; the blue circles represent the state estimates $\hat{x}$. The experimental results show that the estimated displacements are generally close to the actual value indicating a good regression performance. Furthermore, Table~\ref{wake_est_result} reveals that a larger longitudinal displacement leads to a slightly increased root mean square error (RMSE) at 0.07 compared to that of 0.05 from the smaller longitudinal displacement. Meanwhile, the standard deviation of the estimation error from both cases are satisfactorily small at 0.0042 and 0.0012, respectively.

Figures~\ref{fig: result_est}(c)-\ref{fig: result_est}(f) show the estimation results regarding the traveling speed ${v}$ and the direction of motion ${d}$ of the leader propeller represented by the confusion matrices that demonstrate the classification performance. The rows and the columns of the confusion matrix correspond to the actual and predicted classes, respectively. The diagonal and off-diagonal elements correspond to the true and false observation results, respectively.
Figs.~\ref{fig: result_est}(c) and \ref{fig: result_est}(d) illustrate the estimation performance regarding the direction of motion ${d}$. $P$ and $N$ represent along the positive and negative $y$-axis directions, respectively. It is observed that the accuracy of the direction state classification is greater than 99.5\% for the smaller longitudinal displacement $y=250$ mm while the accuracy is greater than 99\% for the larger longtudinal displacement $y=300$ mm. The RMSE is less than 0.2\% for both the cases. 
Specifically, Figs.~\ref{fig: result_est}(e) and \ref{fig: result_est}(f) illustrate the estimation performance regarding the traveling speed ${v}$ where 5 different lateral speeds of $0.4,0.5,0.6,0.7,0.8$ m/min are used corresponding to the first to the fifth columns of the confusion matrix labeled as $1$ to $5$. We observe that the accuracy of the lateral speed classification is consistently greater than 91\% with a mean square error less than 1.5\%.

\begin{table*}[h]
    \caption{The experimental results of the statistical estimation performance for all the three motion states in Case 1 and Case 2 using both optimized and random task weights.}
    \centering
    \begin{tabular}{p{4.5cm}p{3cm}p{3cm}p{3cm}p{1.5cm}}
        \hline
        Case & RMSE of $\hat{x}$ & ACC$_1$ of $\hat{v}$ & ACC$_2$ of $\hat{d}$ & Fitness $g$\\
        \hline
       Case 1 (w. OPT Params)              & 0.0498$\pm$ 0.0042  & 97.37$\pm$0.72\% & 99.75$\pm$0.11\%    & 0.07849          \\
        Case 1 (w. RND Params)              & 0.0661$\pm$ 0.0120    & 93.47$\pm$7.32\%        & 98.31$\pm$1.86\%  & 0.18397     \\
        Case 2 (w. OPT Params)      & 0.0712$\pm$ 0.0012     & 95.58$\pm$1.21\%       & 99.37$\pm$0.21\%    & 0.12165   \\
       Case 2 (w. RND Params)      & 0.0645$\pm$ 0.0159     & 89.16$\pm$24.23\%       & 98.89$\pm$1.9\%    & 0.1477   \\
        \hline       
    \end{tabular}
    \label{wake_est_result}
\end{table*} 
The satisfactory experimental results validate the proposed motion state estimation algorithm, which further indicates that the propeller wake features extracted from the ALL distributed pressure measurements are sufficiently rich for estimating the motion states of a leader propeller-driven underwater robot in a leader-follower formation.

\subsection{Comparison Study}
To demonstrate the effectiveness of our proposed estimation algorithm, we conducted a comparative experiment, comparing our approach with two traditional flow sensing methods from the literature including the frequency analysis-based method \cite{Qiu2023TASE,Liu2022SNA} and the learning-based method \cite{Xu2023POF,Xu2022BiB,Rodwell2023BiB}. Specifically, the frequency analysis-based method employs short-time Fourier transform (STFT) to extract localized time-varying frequency features from pressure measurement time series. These features are then inputted into a two-dimensional convolutional neural network to generate estimated states of relevant wake flows. The learning-based method adopts the one-dimensional convolutional neural network (1DCNN) to directly train upon the raw pressure measurement time series.

Figure~\ref{fig: compare} shows the experimental results of the statistical estimation accuracy regarding the three motion states of interest. The comparison reveals that our proposed flow estimation method consistently outperforms the conventional flow sensing algorithms across different estimation states and experimental conditions. Specifically, RMSE of the estimated lateral displacement $\hat{x}$ decreases by 75\% and {30}\%, compared to the conventional frequency analysis-based and learning-based methods, respectively. Similarly, ACC of the estimated traveling speed $\hat{v}$ increases by 68\% and {45}\%, and ACC of the estimated direction of motion $\hat{d}$ increases by 49\% and {27}\%, compared to frequency analysis-based and learning-based methods, respectively. These results illustrate the significant advantage of the proposed method over conventional flow sensing approaches in addressing the unique propeller wake estimation challenge.

\section{Analysis and Discussion}
\subsection{Length of Time Series on Estimation Performance}
Time series of the ALL pressure measurements are used to capture the dynamics of the propeller wake in this paper. Intuitively, the sequence length of the time series is critical in the extraction of the dynamic flow features. Studies are conducted to further investigate the influence of the sequence length on the estimation results, aiming to provide insights into the relationship between required data length and the motion states of the leader propeller. Experimental data collected at longitudinal displacement $y=250$~mm is used while all the other parameters are kept the same as in aforementioned experiments. The sequence lengths of the pressure measurement data for comparative analysis in the estimation algorithm are selected to be $sl = [32, 48, 64, 80]$.

\begin{figure}[!htb]
    \centering
   \includegraphics[width=0.9\linewidth]{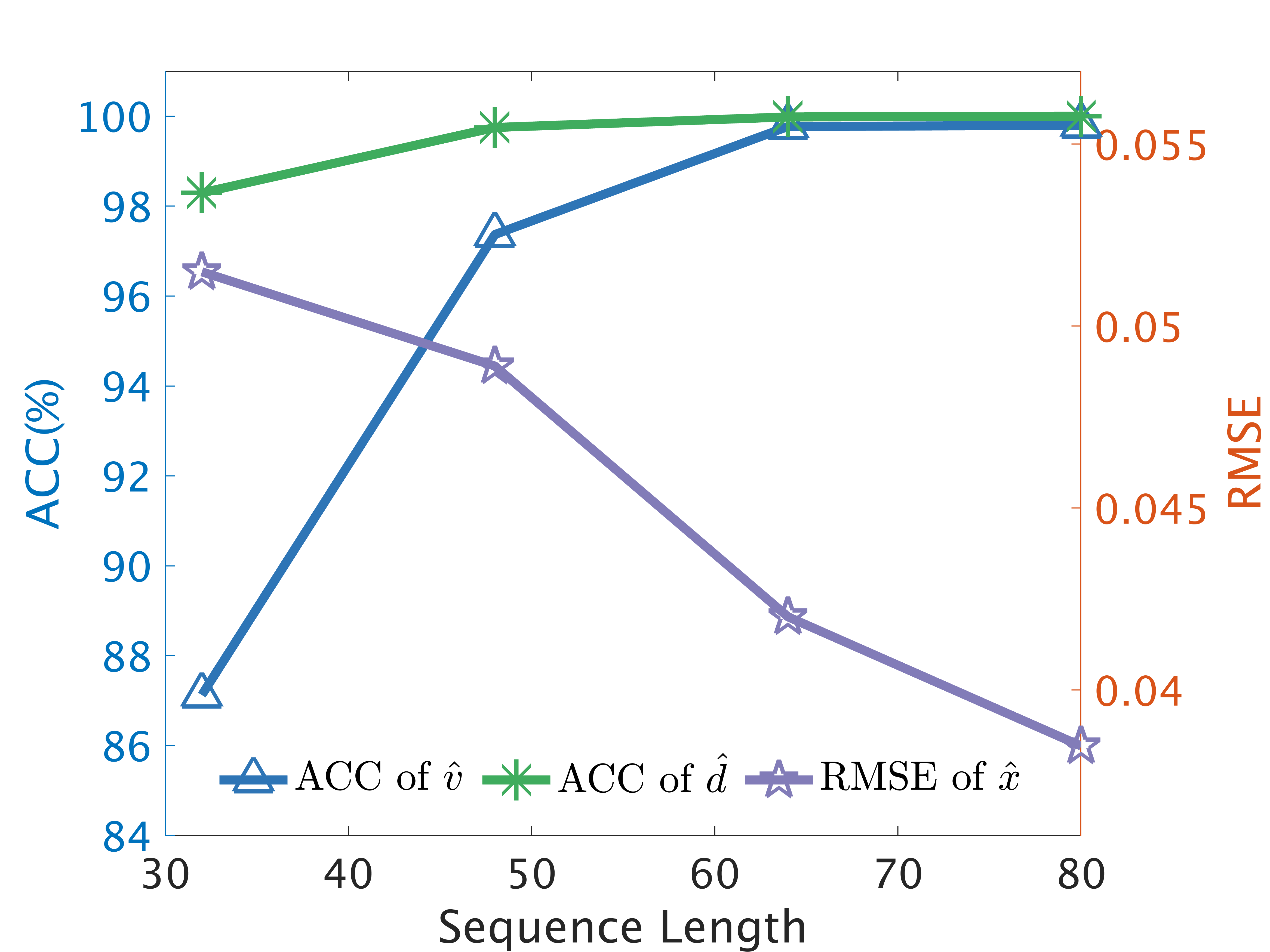}
    \caption{The experimental results of the motion state estimation performance with respect to the sequence length of the time series used in the estimation algorithm. Specifically, the estimation performance ACC of $v$ and $d$ uses the axis scale on the left, and RMSE of $x$ uses the axis scale on the right.}
    \label{fig: result_squence}
\end{figure}

Figure~\ref{fig: result_squence} shows the experimental results of the influence of the time-series sequence length on the estimation performance represented by the corresponding evaluation metrics. 
We observe that as the sequence length increases, the RMSE of the lateral displacement decreases slightly indicating a gentle performance improvement. On the other hand, the classification accuracy of both the traveling speed and the motion of direction increases considerably when the sequence length increases. Overall, the longer length of the time-series is used in the estimation algorithm, the higher the motion estimation performance is achieved, which we conjecture comes from the extra flow dynamics information embedded in the added sampled data. However, longer sequence length leads to more time delay in the estimation process, thus possibly hindering real-time control tasks.
Comprehensively considering the estimation performance and the real-time requirement of the ALL sensing system, the selection of the sequence length of the sensor measurement time series is expected to be resolved with respect to different applications. Particularly, $sl = 64$ is adopted in this paper.

\subsection{Optimized Task Weights on Estimation Performance}

We conducted a comparative experiment using WOA-based optimized weights and randomly selected weights to assess their influence on overall estimation performance. In the comparative experiment, we trained 10 estimation models based on 10 sets of randomly generated task weights and another 10 estimation models based on the WOA optimized weights. We compared and analyzed the averaged fitness value $\bar{g}$ over 10 estimation models and the statistical state estimation accuracy between using random and optimized weights.

Table~\ref{wake_est_result} illustrates the experimental results. We observe a significant decrease in fitness $g$ using optimized weights compared to using random weight, by roughly 57\% and 17\% in Case 1 and Case 2, respectively.
Furthermore, in Case 1, we notice a significant enhancement in estimation performance with optimized weights. Specifically, there is an improvement in estimation accuracy of 4\% in speed $v$, 1.4\% in direction of motion $d$, and 24.6\% in displacement $x$. 
However, the estimation accuracy for displacement $x$ in Case 2 decreases by 0.6\%, which we conjecture is attributed to the highly-coupled nature of multi-objective optimization.
In addition, the standard deviations of the estimation accuracy for all three states are reduced significantly by at least 65\% with optimized weights, indicating a more robust and consistent estimation performance across all motion states and testing conditions. Therefore, optimizing the algorithm to obtain optimization weights is crucial.

\subsection{WOA vs GA and PSO}
\begin{figure}[htp!]
    \centering
    \includegraphics[width=0.95\linewidth]{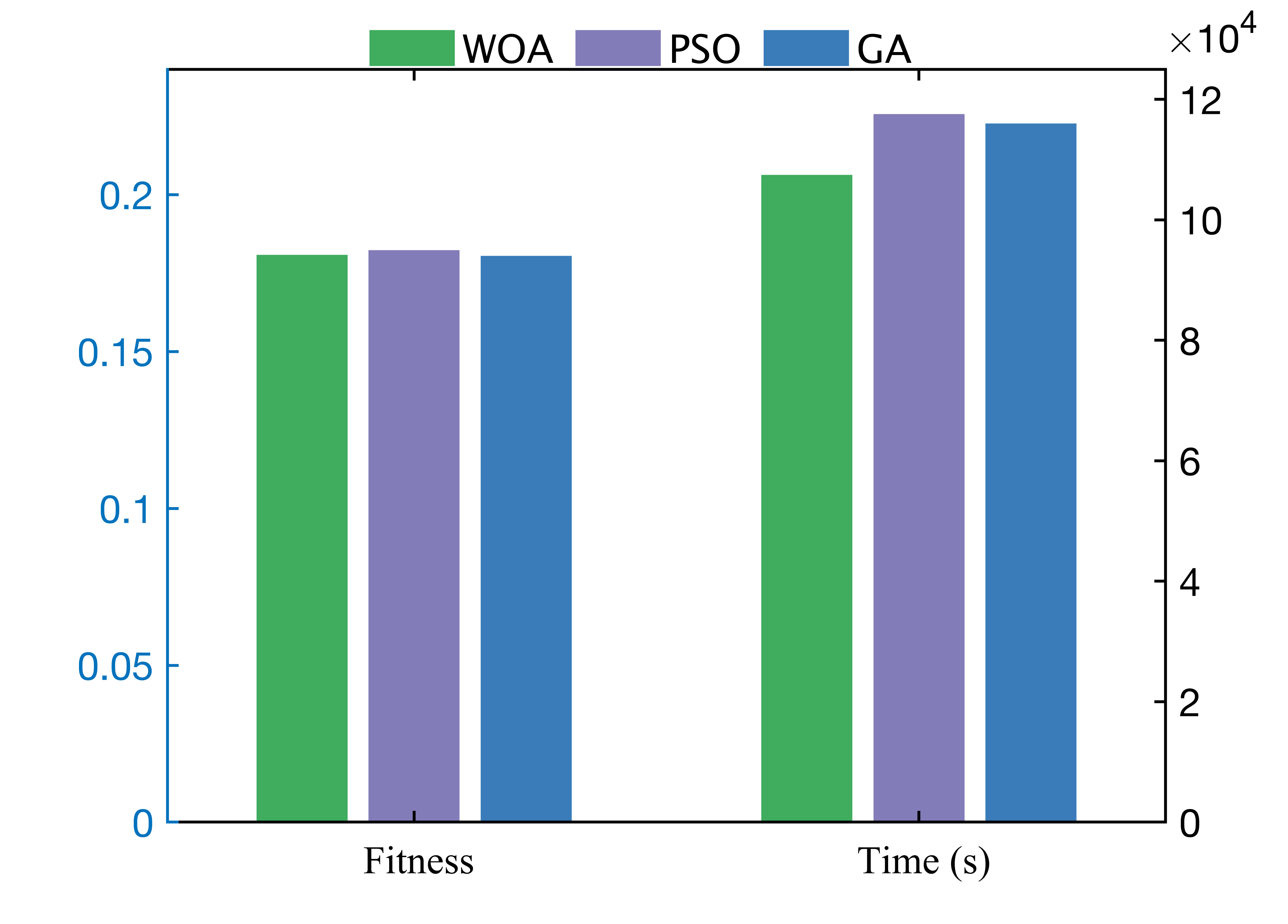}
    \caption{Experimental results of the averaged fitness value $\bar{g}$ and the computational time consumption, comparing WOA with GA and PSO in estimation model training. While WOA exhibits similar fitness value, it significantly reduces computational time, thus making WOA a preferred choice for optimizing network parameters in this paper.}
    \label{fig:woa_compare}
\end{figure}

To justify our choice of the optimization algorithm, we conducted a comparative experiment, comparing WOA with classical Genetic Algorithm (GA) and Particle Swarm Optimization (PSO) algorithms. We adjusted some hyperparameters to expedite the process, e.g., reducing training steps to 100 from 200 and batch size to 512 from 64. For each optimization algorithm, we conducted five repeated tests varying only the initial weight values while keeping all other optimizer parameters unchanged. Fig.~\ref{fig:woa_compare} shows the experimental results of averaged fitness value and computational time over the five repeated tests. While the fitness value of WOA is approximately 0.2\% lower than that of GA and 0.03\% higher than that of PSO, the WOA algorithm significantly reduces optimization time by approximately 7\% and 8\% over GA and PSO, respectively.
Comprehensively considering the network complexity and extended computational time, we selected WOA for its effectiveness and efficiency as the task weight optimizer in this paper.

\begin{table}[htbp]
\small
	\centering
	\caption{The comparison between our method and other selected methods of ALL-based state estimation of underwater robots.}
	\label{ablation}
	\begin{tabular}{c|c|c}
		\hline
		  & \tabincell{c}{wake flow generation}  &  \tabincell{c}{multi-state\\estimation} \\
           \hline
		 \tabincell{c}{ MIT \cite{Gao2018JFM} } & \tabincell{c}{fishlike body undulation \\+ caudal fin flapping}  & \XSolidBrush       \\
		\hline
		 
		 \tabincell{c}{PKU  \cite{Wang2015IROS,Zheng2017BiB,Zheng2019SMC,Li2020NC} } & robot fish tail flapping  & \XSolidBrush      \\
            \hline
            \tabincell{c}{NTU \cite{Yen2022BiB}}  & robot fish tail flapping  & \CheckmarkBold     \\
           \hline
		 \tabincell{c}{HEU \cite{Xu2022BiB}}  & robot fish body undulation   & \XSolidBrush       \\ 
         \hline
          \tabincell{c}{UMD \cite{Free2018BiB}}  & Joukowski hydrofoil flapping   & \CheckmarkBold        \\ 
         \hline
            Ours & high-speed rotating propller & \CheckmarkBold     \\
		\hline
	\end{tabular}
\end{table}

\subsection{Uniqueness in Propeller Wake Sensing}
Table~\ref{ablation} presents a comparison between our proposed method and others from the literature, focusing on the generation method of the flow field and whether multiple motion states are simultaneously estimated. Existing studies primarily address flow fields generated by periodic body undulation and tail flapping such as in bioinspired robotic fish, while our study deals with the dynamic and complex wake flow generated by high-speed rotating propellers. Moreover, our method enables simultaneous estimation of multiple motion states through a multi-output network design combining regression and classification operations. Therefore, we consider the problem addressed in this paper unique, significantly different from the existing studies.

\subsection{Limitations and Mitigation Measures} 
We would like to briefly discuss the limitations of the proposed propeller wake sensing method and explore mitigation measures to enhance its real-world application.
First, the data-driven algorithm relies on pre-collected training data, so conducting more experiments is necessary to expand the dataset and cover more types of complex motions. Secondly, the horizontal sensor arrangement limits the sensing of motion states to the corresponding plane. Therefore, enhancing ALL with more distributed pressure sensors is critical for improving its spatial sensing capabilities. Thirdly, in this study, ALL was fixed on the experimental platform without considering sensor noise and pressure perturbation caused by its motion. A viable solution is to integrate inertia measurement units and fuse sensor measurements to mitigate such influences.

\section{Conclusions}
This paper investigated the challenging problem of estimating the relative motion states of a propeller-driven leader underwater robot in the leader-follower formation. Making a hypothesis that the ALL pressure measurements provide partial but sufficiently rich information of the propeller wake of a leader underwater robot, we proposed a novel motion state estimation method that assimilated distributed pressure measurements sampled from the dynamic and complex propeller wake. Specifically, to estimate the lateral displacement, speed, and direction of motion of the leader propeller, a hybrid network that combined 1DCNN and BiLSTM was designed to extract the spatiotemporal characteristics of the flow field, followed by a multi-output network for state estimation. An experimental testbed was constructed. Extensive experimental tests were conducted, the results of which validated the proposed estimation method. 

\bibliographystyle{IEEEtran}
\bibliography{main}

\end{document}